\documentstyle[12pt,mkstyle,twoside]{article}
\title{What Does a Belief Function Believe In ?}

\newcommand{\Bem}[1]{}

\newcommand{\V}{\mbox{\bf V}}
\newcommand{\SS}{\mbox{\bf S}}

\newcommand{\LitStelle}[1]{  

\vspace{-2.5mm}

\bibitem{#1}  }
\Bem{

}

\author{Andrzej Matuszewski, Mieczys{\l}aw A. K{\l}opotek} %

\begin{document}

\machetitel







\begin{center}
ABSTRACT 
\end{center}

The conditioning in the Dempster-Shafer Theory of Evidence 
has been defined (by Shafer \cite{Shafer:90} as combination of a belief
function and of an "event"
via Dempster rule. 
On the other hand Shafer \cite{Shafer:90} gives a "probabilistic" 
interpretation of a belief function (hence indirectly its derivation from a
sample). Given the fact that  conditional probability distribution
of a sample-derived probability distribution
is a probability distribution derived from a subsample (selected on the
grounds of a conditioning event), the paper investigates the empirical
nature of the Dempster-
rule of combination. It is demonstrated that the so-called "conditional"
belief function is not a belief function given an event but rather a belief
function given manipulation of original empirical data.\\
Given this, an interpretation of belief function different from that of Shafer
is proposed. Algorithms for construction of belief networks from data are
derived for this interpretation.\\

\Bem{
\begin{center}
O CZYM JEST PRZEKONANA FUNKCJA PRZEKONANIA
\end{center}

W artykule badana jest empiryczna natura regu{\l}y Dempstera sk{\l}adania 
niezale\.{z}nych funkcji przekonania w Matematycznej Teorii Ewidencji. Pokazuje
si\c{e}, \.{z}e tzw. "warunkowa" funkcja przekonania nie jest funkcj\c{a} przekonania
pod warunkiem zaj\'{s}cia okre\'{s}lonego zdarzenia lecz raczej funkcj\c{a} przekonania
pod warunkiem manipulowania danymi empirycznymi.\\
Dlatego proponuje si\c{e} odmienn\c{a} of shaferowskiej \cite{Shafer:90}
interpretacj\c{e} funkcji przekonania. Prezentowane s\c{a} algorytmy do konstrukcji
sieci przekona\'{n} z danych dla proponowanej nowej interpretacji funkcji
przekonania.\\

\newpage


\thispagestyle{empty}

\quad

\newpage

}

\section{Introduction}

The Dempster-Shafer (DS) Theory (DST) or the Theory of Evidence 
\cite{Shafer:76}, 
\cite{Dempster:67}  is  considered  by  many  researchers  as   an 
appropriate tool to 
represent various aspects of human dealing with uncertain knowledge, 
especially for representation of partial ignorance \cite{Smets:92}, though 
this view has been challenged by various authors (compare the 
presentations and discussions in International Journal of 
Approximate Reasoning (IJAR), special issues in  Vol. 1990:4 No. 5/6 and 
Vol. 1992:6 No.3; see also \cite{Halpern:92}, \cite{Fagin:91}.

This paper is intended to shed some light onto the dispute over adequacy of 
the DST from the technical point of view. The authors of this paper 
have been  engaged 
 in a project having as its goal the implementation of an expert system 
dealing with uncertainty via DST methodology mixed with the Bayesian approach
\cite{Klopotek:92}. 
 Knowledge is represented by a belief network to enable application of the 
reasoning system based on the work of Shenoy and Shafer \cite{Shenoy:90} 
(Shenoy-Shafer's axiomatic  framework  encompasses  both  DST  and 
Bayesian-like 
reasoning scheme). It has been an ultimate goal of the developers of that 
expert system to support also knowledge acquisition from data.  Literature 
provides with many methods of recovery of Bayesian 
belief network from data (and 
in certain cases from additional appropriate hints of 
an expert), if the belief network should be a tree \cite{Chow:68}, a polytree 
\cite{Rebane:89}, or a general-type (usually sparse)  network 
\cite{Cooper:92}, \cite{Spirtes:90b}  

  As "generalized probability" is a term frequently used to characterize the 
DS 
belief function, it seems plausible to try to generalize Bayesian  methods  
onto recovery   of   Dempster-Shafer   belief 
networks from data. However, as the above-mentioned discussion in IJAR 
demonstrates, the relationship between  empirical
 frequencies and DS belief  functions 
seems to be far from being clear. 

We agree with Smets \cite{Smets:92} that fundamental deviation 
of DST from any probabilistic measure of uncertainty lies  in  the 
DS rule  of  combination  ($\oplus$)  which  serves  as  a  way  of 
conditioning the  overall  DS  distribution  on  some  event  (see 
\cite{Shafer:90}). 

In this paper we assume that DST notions like basic probability assignment 
or mass function $m$, belief function $Bel$, 
pseudo-mass and pseudo-belief functions,
combination $\oplus$,  
marginalization $\downarrow$ and empty extension $\uparrow$ operations are 
well understood \cite{Shafer:76}, \cite{Shenoy:90}.

\section{Nature of Conditioning in DST}

Let us consider for a moment how "empirical" conditioning may be 
viewed in the probability theory. Let a  probability  distribution 
be defined as relative frequency over a (large) population. Let us 
want to condition on an event, say $\{\omega | \alpha(\omega)\}$ 
where $\alpha$ is a predicate - a logical expression  in  variables 
describing the population. Then we select all the objects $\omega$ 
of the population which match the predicate  $\alpha(\omega)$  and 
the conditional distribution $P(.|\alpha)$ will  be  the  relative 
frequencies within this subpopulation.

Let us try  to  do  the  same  with  DS  Bel's.  Following  Shafer 
\cite{Shafer:90} we may be tempted to interpret a  valuation of  an 
object $\omega$ of a population with  a set  A  as  a 
statement that our variable of interest takes for this object  one of 
the values mentioned in A, but we do not know which  one  of  them 
(and are not ready even to select any proper subset of A). A  will 
be treated as our most specific commitment to  the  value  of  the 
variable.  Under  these  circumstances,  the   basic   probability 
assignment  function  m  may  be  understood  as  the  probability 
distribution (read:  relative  frequencies)  of  such  commitments 
within  our  population.  This  is,  in  fact,  the  way  as   the 
"generalized probability" in  \cite{Halpern:92},  or  families  of 
probability distributions in \cite{Kyburg:87} may  be  understood. 
Now let us go over to conditioning on an event 
 $\{\omega | \alpha(\omega)\}$
 (after \cite{Shafer:90}). It is the beauty of the DST that  there 
exists always a set B that corresponds exactly to such an event. 
If $Bel_B$ is the simple support function capturing  the  evidence 
of the set B (that is $m_B(B)=1$, $m_B(A)=0$ for any $A \neq B$, 
\cite{Shafer:90}) then by definition $Bel(.|B)=Bel \oplus 
Bel_B$
is the belief of $Bel$ conditioned on the event $B$ \cite{Shafer:90}. 
 But how does this definition "run" on a population of objects~? It 
has been demonstrated \cite{Klopotek:93} that we can view  it  the 
following way: We  take  the  predicate  $\alpha$  and  check  the 
population object by object. Some of them deny the  predicate.  We 
reject them as the frequentist model  of  conditional  probability 
does. Some of them meet the predicate - and we select them for our 
subpopulation - as conditional probability model does.  But  there 
are some objects for  which  we  are  actually  unable  to  decide 
whether they meet the predicate or not (as our commitment  is  not 
specific enough). And what do we have to do in this case  to  meet 
numerically the DS rule of combination~? We have to accept them~! 
But (Alas!) this is not enough. We have to change our commitment - 
we have to make our  commitment  for  a  particular  object   more 
specific so that it meets the predicate $\alpha$. So even  if  our 
commitment to the value of the attribute for this object may  have 
been correct prior  to  conditioning  (that  is  the  variable  of 
interest took  for  the object one of the values mentioned in  the 
prior commitment), it may be not correct after  the  conditioning. 
That is, after the conditioning we work with a subpopulation  with 
partially incorrect valuations (not corresponding  with  empirical 
reality), and we combine evidence further ..... . (Classical!) 
probability theory does not do things  like 
this - it assumes that measuring sequence has  no  impact  on  the 
value of a variable (Of course, there are several   non-classical 
probability  theories  which  took  into  account  possibility  of 
disagreement between  various  observations  if  the  sequence  of 
making observations is changed, but obviously  most  frequentist 
interpretations of DST didn't consider them). 

We feel that  this (essentially numerical) argument explains 
most  of  apparent  contradictions  derived   from  frequentist    
interpretations of the DST. But the DST will not be helped with if 
left with the impression of telling us  lies about the  population 
(we have to say, plausible lies, because by definition  we are 
unable to check by observation 
if strengthening of a commitment for an  object  is  in  fact 
correct or not, because if we were  able  to  carry  out  such  an 
observation then  our  prior  commitment  would  have  been  more 
specific). So instead of saying that the variable  takes  for  the 
object one of the values in A, we could say that the  variable  is 
set-valued and takes for the object all  the  values  in  A  (this 
would be a kind  of  random  set  interpretation
\cite{Nguyen:78}).  In  this  case 
conditioning could be viewed as rejecting some of  the  values  of 
the object which are not of interest. Then after  conditioning 
the object would have a commitment corresponding to the values  it 
really takes, though some other values  were  ignored  as  not  of 
interest. However, this view would contradict the usage  of  Bel's 
to represent material implication $P(\omega) \rightarrow Q(\omega)$ 
in form of a set $\{(P(\omega),Q(\omega)), 
(\lnot P(\omega),Q(\omega)), 
(\lnot P(\omega),\lnot Q(\omega))\}$ because it would lead to  the 
impression that at the same time  $P(\omega) \land Q(\omega)$ and 
 $\lnot P(\omega) \land Q(\omega)$ hold which is  counterintuitive 
as  $P(\omega) \land \lnot P(\omega)=FALSE$.\\

\section{An Alternative View of DST}

Hence an alternative view  of  DST  is  required.  One  has  been 
developed in \cite{Klopotek:93}. We present  it  here  informally. 
Instead of saying that the set A expresses that "the  variable  of 
interest takes one of the values in  A"  as  well  as  instead  of 
saying  
 that the set A expresses that "the  variable  of 
interest takes all of the values in A" a compromise  is  proposed: 
it is assumed that the variable cannot be observed  directly,  but 
only via some measurement procedure (with some special  properties 
ensuring consistence),  and 
 the set A expresses that "the measurement procedure yielded  TRUE 
when testing if $X=a_i$ ($X$ - the variable of interest) 
for all the $a_i \in A$ and for no $a_i \not\in A$".  If  we  make 
conditioning, the objects are labeled, and the measurement  method 
takes into account the labeling  of  objects  by  refraining  from 
carrying out tests on variable values outside  of  the  label.  In 
this way the following is achieved:
\begin{itemize}
\item  before and after every conditioning the  interpretation  of 
the commitment A is the same for not rejected objects: 
 "the measurement procedure yielded  TRUE 
when testing if $X=a_i$ ($X$ - the variable of interest) 
for all the $a_i \in A$ and for no $a_i \not\in A$". 
\item the impact of conditioning onto measurement results is taken 
into account - via labeling
\item  any  logical  contradictions  resulting  from  random   set 
interpretation are avoided: we do not say "$X$  takes  the  value" 
but that "$X$  has  been  measured  to  be",  and  contradictions 
resolve in imprecision of measurement method.
\end{itemize}

The  importance  of  such  an  interpretation   is   not   to   be 
underestimated: a way is paved  towards  experimental  studies  of 
populations with DS belief distributions.\\

To demonstrate this a development of a method 
of a tree/polytree factorization of a joint DS belief distribution 
for purposes of Shenoy/Shafer uncertainty propagation \cite{Shenoy:90}  is 
briefly outlined.

We define {\em mk-conditional belief function}   
$Bel ^{X | X_i}(A)$ as any pseudo-belief function solving the equation 
 $Bel=Bel ^{\downarrow X_i} \oplus  Bel ^{X | X_i}$
Notice, that in general this equation has no unique solution, and a solution 
being a proper belief function does not  always exist.

 A 
{\em DS Belief 
 network}   be  \cite{Klopotek:93d} a pair (D,Bel) where D is a dag (directed 
acyclic graph) and Bel  is a DS belief 
distribution called the {\em underlying distribution}. Each node $i$ in D 
corresponds to a variable $X_i$  in Bel, a set of nodes I corresponds to a 
set of variables $X_I$ and $x_i, x_I$
 denote values drawn from the domain of $X_i$ 
 and from the (cross product) domain of $X_I$ respectively. Each node in the 
network  is regarded as a storage cell for any  distribution 
$Bel ^{\downarrow \{X_i\} \cup X_{\pi (i)} |  X_{\pi (i)} }$
 where $X_{\pi (i)}$ is a set of nodes corresponding to 
the 
parent nodes $\pi(i)$ of $i$.  The underlying distribution represented by a 
DS belief network is computed via:
$$Bel  = \bigoplus_{i=1}^{n}Bel ^{\downarrow \{X_i\} \cup X_{\pi (i)} |  
X_{\pi (i)} } $$

\Bem{
\begin{df}  \cite{Geiger:90} 
A {\em trail } in a dag is a sequence of links that form a path in the 
 underlying undirected graph. A node $\beta$ is called a {\em head-to-head 
node}  with 
respect to a trail t if there are two consecutive links $\alpha \rightarrow 
\beta$ and $\beta \leftarrow \gamma$ on that t. 
\end{df}

\begin{df}  \cite{Geiger:90} 
A trail t connecting nodes $\alpha$ and $\beta$ is said to be {\em active } 
given a set of nodes L, if (1) every head-to-head-node wrt t either is or has 
a descendent in L and (2) every other node on t is outside L. Otherwise t is 
said to be {\em blocked } (given L).
\end{df}

\begin{df}  \cite{Geiger:90} 
If J,K and L are three disjoint sets of nodes in a dag D, then L is said to 
{\em d-separate } J from K, denoted $I(J,K|L)_D$  iff no trail between a node 
in J and a node in K is active given L.
\end{df}
}
Let, after \cite{Geiger:90}  $I(J,K|L)_D$ denote 
{\em d-separation} of  J from K by L in a directed acyclic graph D, where J,K 
and L are three disjoint sets of nodes in this dag D. 
We shall then define  \cite{Klopotek:93d}

If $X_J,X_K,X_L$ are three disjoint sets of variables of a distribution Bel, 
then $X_J,X_K$ are said to be {\em conditionally independent} given $X_L$ 
(denoted $I(X_J,X_K |X_L)_{Bel}$ iff 
 $$Bel ^{\downarrow X_J \cup X_K \cup X_L |  X_L} 
 \oplus Bel ^{\downarrow   X_L } =
 Bel ^{\downarrow X_J  \cup X_L |  X_L} \oplus
 Bel ^{\downarrow X_K \cup X_L |  X_L} 
 \oplus Bel ^{\downarrow   X_L } $$
%
$I(X_J,X_K |X_L)_{Bel}$ is called a {\em 
(conditional independence) statement}

\begin{th} \label{IDIBel} \cite{Klopotek:93d}
Let $Bel_D=\{Bel|$(D,Bel) is a DS belief network\}. Then:
$$I(J,K|L)_D$$ iff $$I(X_J,X_K |X_L)_{Bel}$$ for all $Bel \in Bel_D$.
\end{th}

Many authors have connected causality with
the notion of statistical dependence or non-independence. 
We parallel here  \cite{Spirtes:90b} in formulating the following principles, 
while understanding independence as defined above

Let \V  be a set of random DS variables with a joint 
DS-belief distribution. We say that variables X,Y $\in$\V  are
 {\em directly causally dependent} if and 
 only if there is a causal dependency between X,Y (either the value of X 
influences the value of Y or the value of Y influences the value of X or the 
value of a third variable not in \V  influences the values of both X and Y)
that does not involve any other variable in \V.

{\bf Principle I: }  For all X,Y in \V, X and Y are directly causally 
dependent 
if and only if for every subset \SS  of \V  not containing X or Y, X and Y are
not statistically independent conditional on \SS.\\

We  say that {\em B is directly causally dependent on A} provided that A and 
B 
are causally dependent and the direction of causal influence is from A to 
B.\\

{\bf Principle II: } if A and B are directly causally dependent and B and C 
are directly causally dependent, but A and C are not, then:
B is causally dependent on A, and B is causally dependent on C if and only if 
A and C are statistically dependent conditional on any set of variables 
containing B and not containing A or C.\\

{\bf Principle III: } A directed acyclic graph represents a DS-belief
distribution on the variables that are vertices of the graph if and only if\\
for all vertices X,Y and all sets \SS    of vertices in the graph
(X,Y $\notin$ \SS), \SS  d-separates 
X and Y if and only if X and Y are independent conditional on \SS.\\

\begin{th} \label{iiiDSi} {\rm \cite{Klopotek:93d}}
Let Bel be a DS-belief   distribution represented by an acyclic directed 
graph 
G according to Principle III. Then G is an orientation (G has the undirected 
structure) of the undirected graph U that represents Bel according to 
Principle I.
\end{th}
\begin{th}  \label{iiiDSii}  {\rm \cite{Klopotek:93d}}
Principle III implies Principle II.
\end{th}
\begin{th}   {\rm \cite{Klopotek:93d}}
Let $\Gamma$ be the set of directed 
graphs 
that represent DS-belief   distribution Bel according to Principle III. Then 
$\Gamma$ is also the set of directed graphs obtained from P by Principles I 
and II.
 \end{th}%

\section{Belief Networks from Data under New Interpretation}

Based on these purely theoretical considerations it was tried to develop some 
practical algorithms for recovery of belief network structure from data for 
some limited classes of belief networks. It was started with the most 
successful structures of Bayesian networks: the tree and the polytree 
structures. In these efforts, corresponding Bayesian algorithms were exploited
as general frameworks, though details had to be elaborated anew. It is also 
worth mentioning, that, unlike in probabilistic case, a randomized 
generation of a belief function possessing given 
 belief network structure is not a trivial task due to the data-changing 
nature of DS combination.\\

Let us present briefly these new algorithms:\\

The algorithm of Chow and Liu \cite{Chow:68} for recovery of tree structure of
a probability distribution is well known and has been deeply investigated, so 
we will omit its  description in this paper. To accommodate it for the 
needs 
of DST one needs to introduce a definition of distance between variables. 
Regrettably, no such definition having the nice properties of the Chow and
Liu exists, so a similar one has been elaborated:
Let    $p$ be a mass function and $x$ be a pseudo-mass function. Let 
$f(x;p)= \sum_{A; p(A)>0} p(A) \cdot \ln x(A)$, 
where the assumption is made that natural logarithm of a non-positive number 
is minus infinity. The values of $f$ in variable $x$ with parameter $p$ have
range:
$(-\infty, f(p;p)]$.
 Let $g(x;p)=\frac{f(x;p) }{f(p;p)}$.
 The values of $g$ in variable $x$ with parameter $p$ 
range:
$[1,+\infty)$.
 Let $a(x;p)=e ^{1-g(x;p)}$. 
The values of $a$ in variable $x$ with parameter $p$ 
range:
$[0,1]$

By the ternary joint distribution of the variables  $X_1,X_2$ with background 
 $X_3$ we understand the function:\\
$$m ^{\downarrow X_1 \times X_2[X_3]}=$$
$$=(m ^{\downarrow X_1 \times X_3 | X_3} \oplus
 m ^{\downarrow X_2 \times X_3 | X_3} \oplus m ^{\downarrow X_3})
 ^{\downarrow X_1 \times X_2}$$

By the distance (for use with Chow/Liu algorithm)  
$DEP(X_1,X_2)$ we understand the function:\\
$$DEP0_{DS}(X_1,X_2)=1-
\max(a(m ^{\downarrow X_1}\oplus m ^{\downarrow X_2}; m ^{\downarrow X_1 
\times X_2}), \max_{X_3;X_3 \in V-{X_1,X_2}}$$
$$
 \quad a(m ^{\downarrow X_1 \times 
X_2[X_3]}; m ^{\downarrow X_1 \times X_2})) $$\\
with \V \  being the set of all variables.

For randomly generated tree-like DS belief distributions, if we were working 
directly with these distributions, as expected, the algorithm yielded perfect 
decomposition into the original tree. For random samples generated from such 
distributions, the structure was recovered properly for reasonable sample 
sizes (200 for up to 8 variables). Recovery of the joint distribution was not 
too perfect, as the space of possible value combination is tremendous and 
probably quite large sample sizes would be necessary. It is worth mentioning, 
that even with some departures  from truly tree structure a distribution 
could be obtained which reasonable approximated the original one.\\

A well known algorithm for recovery of polytree from data for probability 
  distributions is that of Pearl \cite{Pearl:88}, \cite{Rebane:89},  
we refrain from describing it 
here.  To accommodate it for usage with DS belief distributions we had to 
change the dependence criterion of two variables given a third one.

$$Criterion(X_1\rightarrow X_3, X_2 \rightarrow X_3) = 
(1- a(m ^{\downarrow X_1 
\times X_2[X_3]}; m ^{\downarrow X_1 \times X_2}))-$$
$$-(1- a(m ^{\downarrow X_1 } \oplus m ^{\downarrow X_2}; 
 m ^{\downarrow X_1 \times X_2}))$$
\\
 If the above function  $Criterion$ is positive, we assume head-to-head 
meeting 
of edges $X_1,X_3$ and $X_2,X_3$. The rest of the algorithm runs as that of 
Pearl.

For randomly generated polytree-like DS belief distributions, if we were 
working 
directly with these distributions, as expected, the algorithm yielded perfect 
decomposition into the original polytree. For random samples generated from 
such 
distributions, the structure was recovered properly only for very 
large sample 
sizes (5000 for 6 variables), with growing sample sizes leading to 
spurious indications of head-to-head meetings not present in the original 
distribution. Recovery of the joint distribution was also not 
too perfect, due to immense size of  space of possible value combinations.

Other distance and dependence measures than those mentioned above have been 
tried but no clear winner could have been decided so far. \\

Though we are still far away from our goal of developing an efficient 
algorithm for recovery of general DS belief network structure from empirical 
data, our efforts demonstrated, that there exists at least one way of 
connecting the formalism of the Dempster-Shafer Theory with frequencies from 
empirical data (though this may not be the one the creators of this theory 
had  in mind).  At the same time 
our view of the nature of the DS belief functions  was shifted from 
traditional frequentist view to one with accepting changing valuation of 
objects while running the conditioning process. This proved  
helpful
    when 
comparing results of reasoning of the inference engine with the empirical 
distribution given by data.
One of the consequences of this changing valuation of data during a reasoning 
process is the crucial difference between probabilistic and DS belief
networks:
 In probabilistic networks the conditioning of a whole distribution on a set 
of 
variables has exactly the same meaning as conditionality contained in a node 
 of a network. That is, if the variable $X_n$ represented by a node $n$ 
depends 
on the set of variables $X_{\pi(n)}$ then if we calculate the conditional 
probability $P(X_n | X_{\pi(n)}$ on a whole network e.g. via Shenoy/Shafer 
 algorithm \cite{Shenoy:90}, then the result will be exactly the same as is 
the 
 valuation attached to the node $n$ of the  network.  The situation is 
entirely 
different in case of DS networks: the Shaferian $Bel(X_n | X_{\pi(n)})$ 
calculated from the overall network is (and usually must be) in general 
distinct from the valuation (mk-conditioning) $Bel ^{\downarrow \{X_n\} \cup 
 X_{\pi(n)}|X_{\pi(n)}}$ we attach to a node of the network. Clearly, the 
Shenoy/Shafer 
uncertainty propagation algorithm \cite{Shenoy:90} is fully unaffected by the 
lack of 
identity between these two notions of conditioning, and in fact a node 
valuation 
neither in probabilistic nor in DS case is required to have anything to 
do with any notion of conditionality. But attachment of conditionality to a 
node of a belief network is important for understanding the contents of a 
belief network which was invented as a means of representing causal 
dependencies \cite{Spirtes:90b}. Our notion of mk-conditionality  $Bel 
^{\downarrow \{X_n\} \cup 
 X_{\pi(n)}|X_{\pi(n)}}$ gives a node in a DS belief network a local meaning: 
it can be estimated from data using only variables engaged, that is $\{X_n\} 
\cup X_{\pi(n)}$. Notably, this does not hold for the general view of belief 
networks (that is without reference to conditionality)  presented 
by Shenoy  
 and Shafer \cite{Shenoy:90}.To verify the validity of valuation of any node 
of a general form hypertree considered in \cite{Shenoy:90} one may be forced 
to consider the entire hypertree at once. 

\newcommand{\ReadingsIn}{G. Shafer, J. Pearl eds: Readings in Uncertain 
Reasoning, (
Morgan Kaufmann Publishers Inc., San Mateo, California, 1990)}

\end{document}